\documentclass{article}

\usepackage[final]{corl_2026}

\title{OdorSpaces: Visual-Olfactory Embodied Navigation in 3D Environments}

\usepackage{graphicx}
\usepackage{booktabs}
\usepackage{array}
\usepackage{multirow}
\usepackage{amsmath}
\usepackage{caption}
\usepackage{wrapfig}
\usepackage{enumitem}
\usepackage{placeins}
\usepackage{float}

\author{
  Changyeon Lee \hspace{3mm} Kelin Yu \hspace{3mm} Zheng Wei \hspace{3mm} Shuna Ni \hspace{3mm} Ruohan Gao\\[0.1cm]
  University of Maryland, College Park\\[0.1cm]
  \texttt{\{cyeonlee, kyu85, zwei1235, shunani, rhgao\}@umd.edu} \\[0.1cm]\url{https://odorspaces.github.io}
}

\hypersetup{
  pdftitle={OdorSpaces: Visual-Olfactory Embodied Navigation in 3D Environments},
  pdfauthor={Changyeon Lee, Kelin Yu, Zheng Wei, Shuna Ni, Ruohan Gao}
}

\begin{document}
\maketitle
\vspace{-18pt}

\begin{abstract}
Olfaction provides critical information about airborne chemical cues, yet it remains largely underexplored in embodied AI, with limited prior machine olfaction work focusing primarily on semantic odor perception. We introduce \emph{OdorSpaces}, a physics-based simulation platform and benchmark for \emph{visual-olfactory embodied navigation}. OdorSpaces augments Habitat-Sim with 3D odor transport and dispersion simulated by Fire Dynamics Simulator (FDS), enabling agents to navigate photorealistic indoor environments using both visual and olfactory observations. The benchmark includes two representative source-seeking scenarios: invisible gas-leak localization and early-stage fire localization. We develop an initial multimodal reinforcement learning policy that integrates visual observations and olfactory measurements over time. Experiments show that the policy outperforms a diverse set of baselines, establishing a strong reference for future research on visual-olfactory embodied navigation.
\vspace{-6pt}
\end{abstract}

\keywords{Visual-Olfactory Learning, Machine Olfaction, Embodied Navigation} 

\section{Introduction}
Olfaction, the sense of smell, enables organisms to perceive the chemical state of their surroundings through airborne cues, 
allowing them to detect spoiled food, burning materials, hazardous gases, and other signals that may be invisible or difficult to distinguish visually. Accordingly, olfaction has attracted substantial interest across biology, cognitive science, and chemistry~\cite{chem, molecular, central, stevenson2013olfactory}, where prior work has examined how airborne chemicals are sensed, neurally and perceptually represented, and related to molecular structure.
Despite its importance, machine olfaction remains largely underexplored compared to vision, audition, and touch. Existing learning-based work has mainly focused on semantic odor perception \cite{sanchezlengeling2019machinelearningscentlearning, doi:10.1126/science.ade4401, ozguroglu2025newyorksmellslarge, feng2025smellnetlargescaledatasetrealworld,kim2026see}, such as odor classification and representation learning. 

However, odor is not only semantic, but also inherently \emph{spatial}. Airborne chemicals are emitted from localized sources and transported by airflow, diffusion, and mixing, forming concentration fields that vary over space and time.
In many real-world scenarios, the goal is therefore not merely to recognize an odor, but to infer its source: identifying an invisible gas leak before it accumulates, locating an early-stage fire before visible smoke appears, or tracing chemical cues toward concealed substances such as explosives or narcotics.
While pre-installed gas detectors can trigger safety alarms, their fixed locations provide only sparse measurements and are not designed for active source seeking. 

Mobile embodied agents offer a natural way to address this gap. By moving through an environment, they can actively sample chemical concentrations over time to infer where an odor originates. Yet most prior work on embodied navigation has focused on visual, auditory, and language-guided perception~\cite{anderson2018vision, chen2020soundspaces, chen2023soundspaces20simulationplatform, listenact}, with little attention to olfactory sensing. This limits agents in safety-critical scenarios where gas leaks, smoke, or hazardous chemical sources may be invisible, occluded, or difficult to localize visually. Olfactory navigation is also challenging: odor measurements provide only local, noisy evidence of a dynamic concentration field, with weak or patchy readings, misleading local peaks, and directional cues that may emerge only over time. Reliable navigation therefore requires integrating odor measurements over time while using vision to traverse the 3D environment. 
We formalize this challenge as \emph{visual-olfactory embodied navigation}, where an agent uses visual observations and gas sensor measurements to navigate toward an odor source in a 3D environment.

\begin{figure*}[!t]
    \centering
    \includegraphics[width=\textwidth]{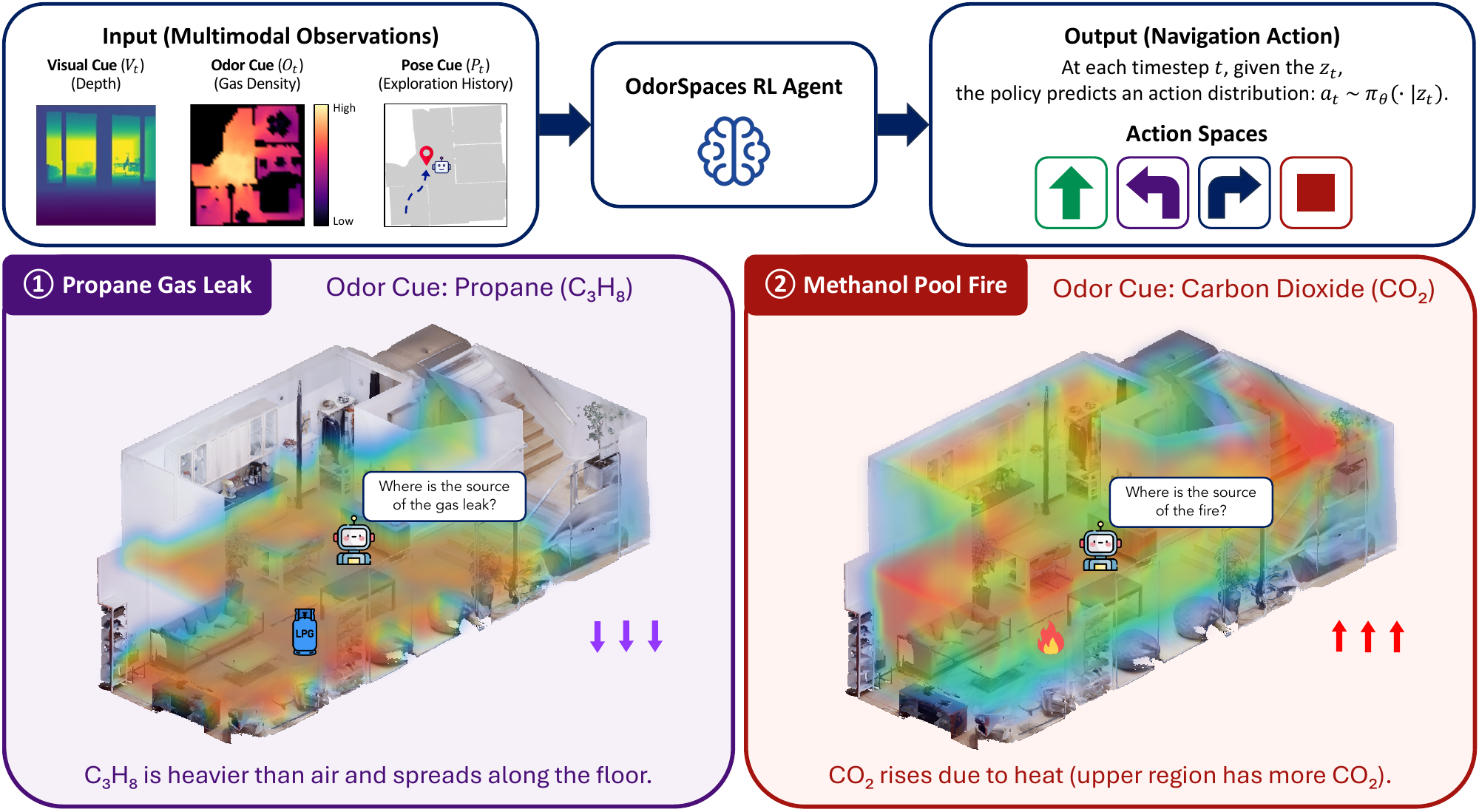}
    \caption{\textbf{OdorSpaces} is a visual-olfactory navigation benchmark where embodied agents navigate toward non-visible odor sources using vision, odor, and pose cues. It uses Fire Dynamics Simulator (FDS) to model physics-based 3D odor fields within Habitat-Sim environments, covering two representative source-seeking scenarios: {propane gas leak and methanol pool fire.}}
    \label{fig:teaser}
    \vspace{-12pt}
\end{figure*}

Training and benchmarking visual-olfactory embodied agents directly in the real world remains challenging. Odor transport is highly sensitive to airflow, temperature, scene geometry, source properties, and ventilation, making odor dynamics hard to reproduce for controlled algorithmic comparison. Prior work often relies on small-scale or geometrically simplified environments \cite{s24072309, MAHMUD2025104915, liu2026scensoryrealtimeroboticolfactory}, which are difficult to scale to visually and geometrically complex indoor spaces. Advancing visual-olfactory embodied navigation therefore requires a simulation framework with physically grounded odor dynamics, a well-defined task, and standardized evaluation protocols in realistic 3D scenes.

We introduce \emph{OdorSpaces}, a physics-based embodied simulation platform and benchmark for visual-olfactory embodied navigation. OdorSpaces uses Fire Dynamics Simulator (FDS) \cite{McGrattan2004} to simulate 3D odor transport and dispersion, and integrates the resulting odor fields into Habitat-Sim \cite{habitat19iccv} environments built from Replica \cite{replica19arxiv} and Matterport3D \cite{Matterport3D} scenes. The benchmark covers two representative real-world source-seeking scenarios: invisible gas leak localization, modeled using
propane (C$_3$H$_8$), a dense gas that settles and spreads near the floor; and early-stage fire localization via combustion-generated carbon dioxide (CO$_2$), modeled as a methanol pool fire where heat-driven buoyancy transports gas upward. 
Together, these scenarios capture distinct physical regimes of odor transport and provide standardized protocols for evaluating odor-guided navigation.

To support this new task, we develop an initial learning-based visual-olfactory navigation policy
that combines visual observations, olfactory measurements, and pose history with recurrent memory for action prediction. Experiments show that the policy achieves competitive performance against diverse baselines, 
highlighting the value of multimodal perception and spatio-temporal modeling for odor-guided navigation. Together, OdorSpaces provides a physics-based simulation platform, a standardized benchmark for visual-olfactory embodied navigation, and a reference baseline for future research on embodied agents with smell sensing.

\vspace{-8pt}
\section{Related Work}
\vspace{-8pt}

\paragraph{Machine Olfaction.}
Odor has long been studied as a chemical and perceptual signal in biology, cognitive science, and chemistry~\cite{chem, molecular, central, stevenson2013olfactory}. Machine olfaction extends this idea to computational systems that sense and interpret odors from chemical measurements, including electronic noses, gas-sensors, and odor recognition systems~\cite{machineolfaction, nose, nosereview}. Recent work has mainly focused on semantic odor understanding, such as modeling structure-to-odor-perception relationships~\cite{sanchezlengeling2019machinelearningscentlearning, doi:10.1126/science.ade4401}, recognizing odor categories from sensor measurements, and learning multimodal odor representations with visual or chemical information~\cite{ozguroglu2025newyorksmellslarge, feng2025smellnetlargescaledatasetrealworld,kim2026see}. These works study what an odor is or how it can be represented, whereas OdorSpaces studies odor as a spatial and temporal signal for embodied navigation.

\vspace{-6pt}

\paragraph{Odor Source Localization.} 
Odor source localization (OSL) studies how agents locate odor sources from local chemical measurements. Classical formulations decompose OSL into plume finding, plume tracking, and source declaration~\cite{s24072309, liu2026scensoryrealtimeroboticolfactory, chen2019odor}. Existing methods include gradient-based search~\cite{russell1995robotic,alt1980biased,jabeen2023robot}, bio-inspired tracking~\cite{ishida1994study,nakata2024olfactory,shigaki2023robust,fukui2025advanced,lochmatter2010bio}, probabilistic or map-based estimation~\cite{vergassola2007infotaxis,rahbar2019design,li2011odor,hutchinson2018unmanned}, and multi-robot search~\cite{marques2006particle,ferri2007explorative,meng2011collective,zhang2015localizing,jain2018odor,gaurav2020single}, and recent learning-based systems use deep RL, recurrent policies, VOC sensor arrays, or vision-language models for odor-guided search~\cite{liu2026scensoryrealtimeroboticolfactory,loisy2023deep,10473576,hassan2024integrating}. While these methods demonstrate the promise of odor-guided navigation, they are often evaluated under simplified plume dynamics, limited scene geometry, and single controlled testbeds. In contrast, OdorSpaces provides a standardized embodied navigation benchmark in photorealistic indoor environments with physics-based 3D odor fields.

\vspace{-6pt}

\paragraph{Multimodal Embodied Navigation.}
Embodied AI has benefited from 3D simulators and benchmarks for navigation, interaction, and policy learning, including Habitat-Sim~\cite{habitat19iccv}, AI2-THOR~\cite{DBLP:journals/corr/abs-1712-05474}, MINOS~\cite{savva2017minosmultimodalindoorsimulator}, and VirtualHome~\cite{DBLP:journals/corr/abs-1806-07011}. Beyond vision-only navigation~\cite{wijmans2019ddppo, chaplot2020object}, multimodal embodied navigation studies how agents use additional sensory cues to perceive and act in 3D environments~\cite{anderson2018vision, chen2020soundspaces, chen2023soundspaces20simulationplatform, listenact, avlen, gao2023sonicverse}. Audio-visual navigation is a representative example, with prior work introducing spatial audio simulation and navigation tasks that use sound as a spatial or semantic cue~\cite{ chen2020soundspaces,chen2023soundspaces20simulationplatform,listenact,chen2021semanticaudiovisualnavigation,chen2021learningsetwaypointsaudiovisual}. OdorSpaces studies a complementary setting: visual-olfactory embodied navigation. Unlike audio, odor is sensed through local gas measurements of a time-varying concentration field shaped by transport, dispersion, accumulation, and scene geometry. As a result, odor cues can be weak, intermittent, and history-dependent, making odor-guided navigation an active sensing problem that requires integrating visual context, gas measurements, and past observations.

\vspace{-6pt}
\section{OdorSpaces: A Benchmark for Visual-Olfactory Embodied Navigation}
\vspace{-6pt}

OdorSpaces couples physically simulated odor fields with photorealistic indoor environments for visual-olfactory embodied navigation. We define the task in Sec.~\ref{sec:task_definition}, describe physics-based odor field simulation in Sec.~\ref{sec:FDS}, the two benchmark scenarios we tackle in Sec.~\ref{ref:scenarios},  Habitat-Sim integration in Sec.~\ref{sec:habitat}, and the agent sensing and action setup in Sec.~\ref{sec:sensing_and_action_setup}.

\vspace{-6pt}
\subsection{Task Definition}\label{sec:task_definition}
\vspace{-3pt}

We define visual-olfactory embodied navigation as an odor-source-seeking task, where an embodied agent uses visual observations and olfactory measurements to navigate toward an odor source. At each decision step $t$, the agent receives an observation $x_t = (V_t, O_t, P_t)$, 
where $V_t$ denotes the egocentric visual observation, $O_t$ denotes local olfactory measurements, and $P_t$ denotes auxiliary pose-history information. Based on $x_t$, the agent selects a navigation action $a_t$ to actively sample the environment and move toward the source. The goal is to reach the odor source within a fixed time budget, reflecting the time-sensitive nature of odor-guided navigation.
In our formulation, the odor source is not visible to the agent, preventing agents from relying on object appearance or visual semantic cues and instead requires active odor sensing. 
Visual observations provide scene context, such as free space, obstacles, traversability, and layout, while olfactory measurements provide local chemical evidence about the surrounding odor field. Pose history provides recent motion context for the policy without revealing the source location. Since each olfactory measurement is local, the agent must actively integrate observations over time during navigation.

\vspace{-6pt}
\subsection{Physics-Based Odor Field Simulation}\label{sec:FDS}
\vspace{-3pt}

We use Fire Dynamics Simulator (FDS)~\cite{mcgrattan2013fds} to generate time-varying 3D gas density fields. FDS is an open-source computational fluid dynamics (CFD) solver developed by the National Institute of Standards and Technology (NIST) for low-speed, thermally driven flows~\cite{mcgrattan2013fds,nist-afm}. This makes it well suited for our benchmark where odor cues are shaped by source properties, buoyancy, density differences, diffusion, mixing, and scene geometry. FDS solves a low-Mach-number form of the Navier--Stokes equations using a large-eddy simulation (LES) formulation on a structured rectilinear grid, with sub-models for turbulence, combustion, radiation, and Lagrangian particle transport~\cite{mcgrattan2013fds}. 
   \begin{wrapfigure}{r}{0.5\textwidth}
    \centering
    \vspace{-4pt}
    \includegraphics[width=\linewidth]{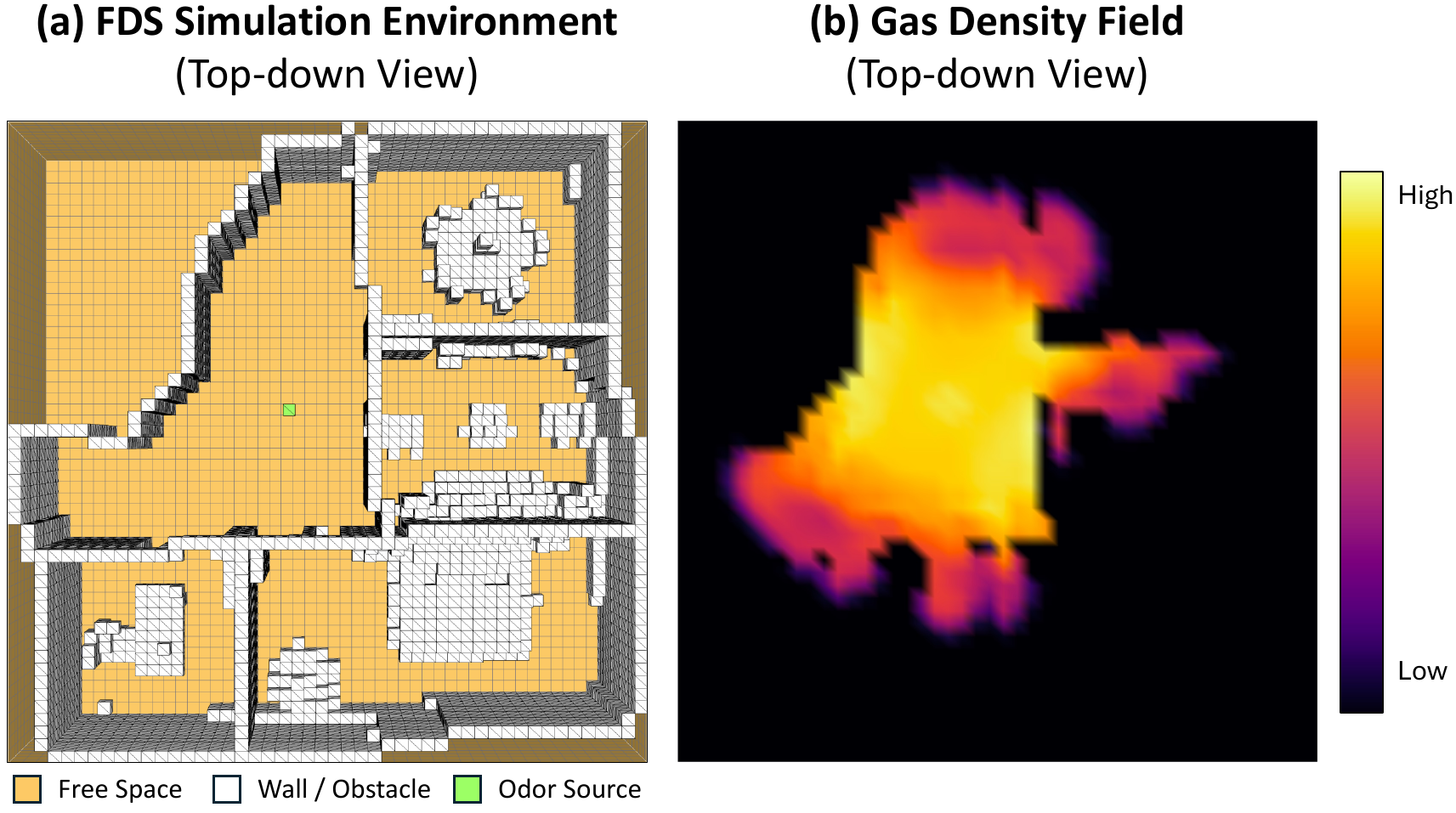}
    \caption{\textbf{Example of the FDS simulation setup.}}
    \vspace{-12pt}
    \label{fig:fds_example}
\end{wrapfigure}
FDS is widely used in performance-based fire safety engineering, including the analyses of complex spaces such as airports, train stations, shopping malls, and stadiums~\cite{nist-afm}. Its predictive capability has been systematically assessed under the ASTM E~1355 framework and documented in the NIST FDS Validation Guide~\cite{mcgrattan2013fdsvalidation}. Prior studies have further validated FDS for smoke and toxic-gas concentrations in compartment fires~\cite{rinne2007vtt}, corridor smoke transport~\cite{floyd2008smoke}, tunnel ventilation dynamics~\cite{tunnelfds}, and atmospheric gas dispersion~\cite{mouilleau2009atmospheric}. Figure~\ref{fig:fds_example} illustrates an example FDS setup, showing the enclosed simulation domain with the odor source and the resulting top-down gas density field.

\vspace{-5pt}
\subsection{Benchmark Scenarios}\label{ref:scenarios}
\vspace{-5pt}

Odor cues can exhibit different spatial structures depending on how they are generated and transported. We therefore instantiate visual-olfactory embodied navigation with two representative, safety-relevant scenarios: \emph{propane gas leak} and \emph{methanol pool fire}.
These scenarios capture distinct source-seeking problems: localizing invisible gas leaks, where dense released gases tend to remain closer to the floor \cite{mynarz2025flammable,ma2026heavygas}, and locating burning materials, where burning odors tend to rise and accumulate in upper regions \cite{sung2021methanol,valoy2023ventilation}. We describe the scenario-specific settings below, with detailed simulation parameters provided in Appendix.

\vspace{-9pt}
\paragraph{{Propane Gas Leak.}} The gas leak scenario models liquefied petroleum gas (LPG) leakage using propane ($\mathrm{C_3H_8}$), the principal component of LPG. Because LPG is typically odorized for human detectability, LPG leakage represents a practical odor-guided source-seeking problem. In our benchmark, we use simulated propane gas density as the olfactory cue for this scenario. Since propane is denser than air, it settles near the floor and forms a low-altitude odor field. Prior CFD work has evaluated FDS for propane dispersion and gas detector analysis \cite{brzezinska2017experimental}. For each source location, propane is released horizontally from the lateral faces of the cubic source object.

\vspace{-9pt}
\paragraph{{Methanol Pool Fire.}} We instantiate the fire-driven source-seeking scenario with a methanol pool fire, a canonical setting in fire modeling and FDS validation studies \cite{mcgrattan2013fdsvalidation,weckman1996experimental}. Combustion-generated carbon dioxide ($\mathrm{CO_2}$) rises through heat-driven buoyancy, and we use its density as a surrogate olfactory cue for fire-related burning odor. For each source location, a methanol pool is placed below the upward-facing face of the cubic source object and ignited using small pilot flames. The pool burns through an FDS pyrolysis model rather than a prescribed heat release rate.

\vspace{-9pt}
\subsection{Integrating Odor-Field Simulation into Habitat-Sim}\label{sec:habitat}
\vspace{-6pt}

We integrate the simulated odor fields with Habitat-Sim \cite{habitat19iccv}, a state-of-the-art physics-enabled embodied AI simulator.
We use 17 real-world scenes from Replica~\cite{replica19arxiv} (9/4/4 for train/val/test) and 25 scenes from Matterport3D~\cite{Matterport3D} (13/6/6 for train/val/test).

Habitat-Sim provides photorealistic egocentric observations, collision checking, and navigation mesh support for embodied policy learning. We discretize each navigation mesh into a 0.5~m 2D navigation grid, which defines the agent's discrete positions and candidate odor source locations. Each FDS odor field is spatially aligned with the corresponding Habitat-Sim coordinate frame, allowing the agent to sample gas density from the same time-varying 3D field while navigating the visual scene. Thus, gas observations are drawn from an aligned spatiotemporal odor field rather than a precomputed static odor map. See Appendix for details.

\vspace{-9pt}
\subsection{Sensing and Action Setup}\label{sec:sensing_and_action_setup}
\vspace{-6pt}

We specify the embodied agent's sensing interface, including visual observation $V_t$, olfactory observation $O_t$, and pose observation $P_t$, followed by the agent's action space.

\vspace{-8pt}
\paragraph{Visual Observation.}
The visual observation $V_t$ is obtained from an egocentric depth camera and captures local scene geometry, including free space, walls, and obstacles. Camera resolution, field of view, and mounting details are provided in Appendix.

\vspace{-8pt}
\paragraph{Olfactory Observation.}
Olfactory observations require specifying where and when gas density is measured. Since embodied olfaction lacks a standardized sensor model, we use a fixed olfactory sensing layout and an idealized sensor that samples continuous-valued gas density from the aligned FDS field. This keeps the benchmark focused on odor-guided navigation rather than hardware-specific sensor design, leaving practical sensor configurations for future work.

At each decision step, the olfactory observation $O_t$ is constructed by sampling gas density across spatial locations and simulation times:
\[
    O_t =
    \left\{
    c(p_t + \Delta_d, h, \tau)
    \mid
    d \in \mathcal{D},\,
    h \in \mathcal{H}_s,\,
    \tau \in \mathcal{T}_s
    \right\},
\]
where $c(\cdot)$ denotes gas density in the aligned FDS field, $p_t$ is the agent position, $\Delta_d$ is the horizontal offset indexed by $d \in \mathcal{D}$, and $h \in \mathcal{H}_s$ is a vertical sensing height, with $\tau \in \mathcal{T}_s$ denoting a sampled simulation time. 
The set $\mathcal{D}$ indexes the agent center and four directional offsets, while $\mathcal{H}_s$ and $\mathcal{T}_s$ specify the sensing heights and temporal sampling window for scenario $s$, respectively. The sampled values are ordered consistently into a fixed-dimensional olfactory vector. Exact sensing locations, heights, and temporal windows are provided in Appendix.

\vspace{-9pt}
\paragraph{Pose Observation.}
The agent receives pose-related observations from GPS and compass sensors. The pose-history observation stores recent position and heading information $P_t = P_{t-n+1:t}$, 
where $n$ is the pose-history window length. Positions are represented on the planar navigation grid, and headings correspond to grid-aligned orientations in the Habitat-Sim coordinate frame. Detailed GPS and compass representations are provided in Appendix.

\vspace{-9pt}
\paragraph{Action Space.}
The agent acts in a discrete navigation action space. The action set consists of \texttt{move\_forward}, \texttt{turn\_left}, \texttt{turn\_right}, and \texttt{stop}. The corresponding grid motion, action duration, and odor-field advancement are provided in Appendix.

\vspace{-6pt}
\section{Visual-Olfactory Navigation Policy}

\begin{figure*}[!t]
    \centering
    \includegraphics[width=\textwidth]{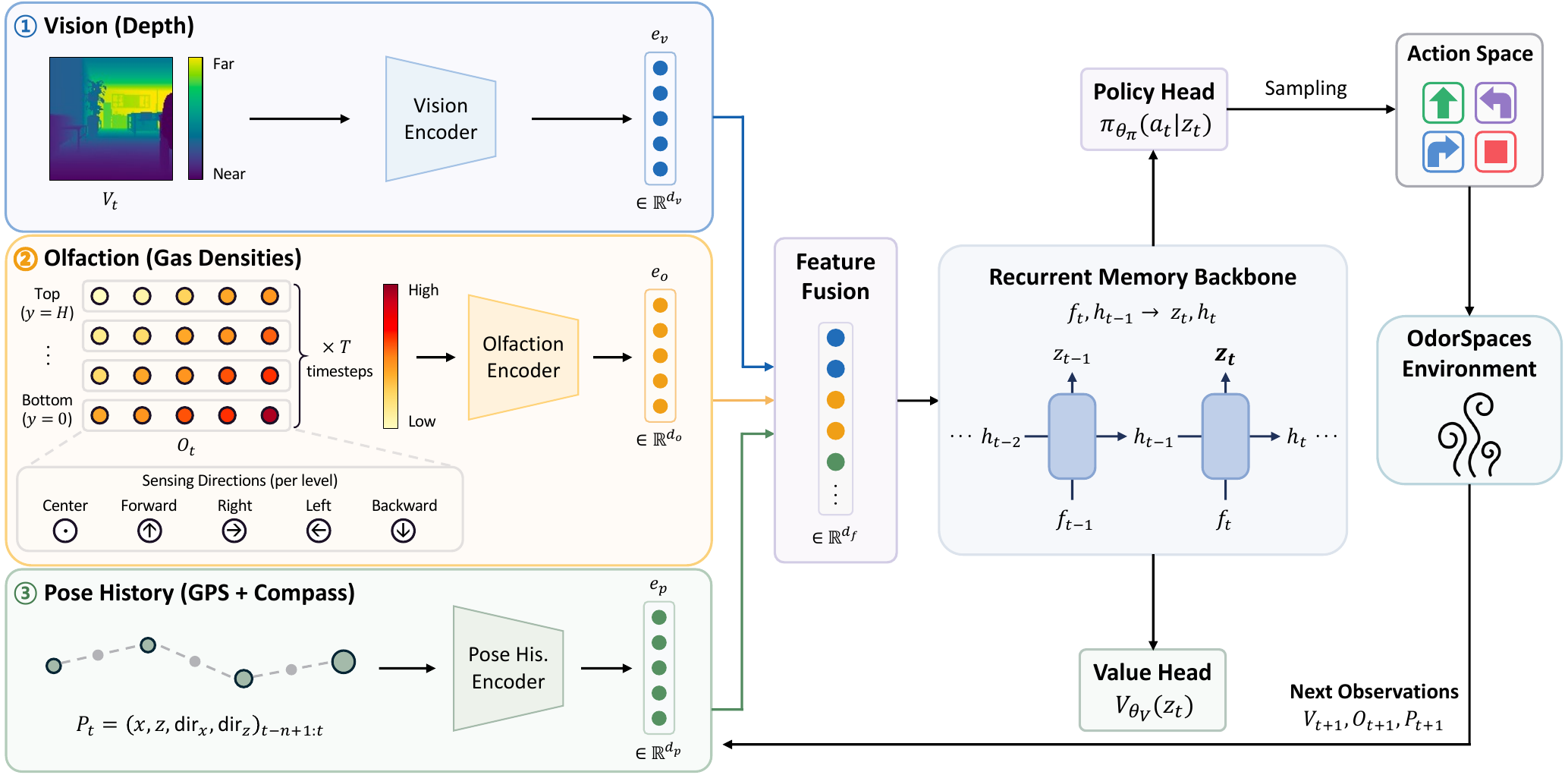}
    \caption{\textbf{Visual-olfactory navigation policy model architecture.}
    We use a multimodal policy that encodes three observation streams: visual observation $V_t$, olfactory observation $O_t$, and pose-history observation $P_t$. The encoded features are fused and passed to a recurrent memory backbone, whose output $z_t$ is used by the policy and value heads for action prediction and value estimation.
    }
    
    \label{fig:policy_architecture}
    \vspace{-9pt}
\end{figure*}

\vspace{-3pt}
\subsection{Architecture}
\vspace{-3pt}

Figure~\ref{fig:policy_architecture} illustrates our multimodal policy for visual-olfactory embodied navigation. Each observation stream is encoded with a modality-specific encoder. The visual observation $V_t$ is processed by a convolutional vision encoder to produce a visual feature $e_v(t)$. The olfactory observation $O_t$ is mapped by a dense olfaction encoder to an olfactory feature $e_o(t)$, and the pose-history observation $P_t$ is mapped by a dense pose-history encoder to a motion-history feature $e_p(t)$. The encoded features are concatenated to form a fused representation:
\[
    f_t = [e_v(t); e_o(t); e_p(t)],
\]
where $[\cdot;\cdot;\cdot]$ denotes  concatenation. The fused feature is passed to a recurrent memory backbone:
\[
    z_t, h_t = \phi_{\mathrm{mem}}(f_t, h_{t-1}),
\]
where $h_t$ is the memory state and $z_t$ is the output feature used for action and value prediction.

The output feature $z_t$ is fed to separate policy and value heads. The policy head outputs a categorical distribution $\pi_{\theta_\pi}(a_t|z_t)$ over the discrete action space, from which an action $a_t$ is sampled during training. The value head estimates the state value $V_{\theta_V}(z_t)$.

\vspace{-6pt}
\subsection{Reward Design}
\vspace{-6pt}
We design the reward using two components: a distance reward with a gas gradient bonus, and a stop reward with a time-dependent bonus. The distance reward encourages the agent to move closer to the source and provides an additional gas gradient bonus when the agent moves toward a location with a higher gas concentration. The stop reward is applied only when the agent selects \texttt{stop}, rewarding successful localization and penalizing incorrect stopping. The reward at decision step $t$ is defined as:
\[
r_t =
\begin{cases}
r_{\mathrm{stop}}, & \text{if } a_t = \texttt{stop},\\
r_{\mathrm{dist}}, & \text{otherwise}.
\end{cases}
\]
For non-stop actions, the distance reward is determined by the change in geodesic distance to the source. For a successful forward movement, we additionally compare the gas concentrations at the previous position $p_{t-1}$ and the new position $p_t$. To isolate the spatial difference in gas concentration from temporal plume variation, both positions are evaluated using the same FDS time window. Let $\mathcal{I}_t$ denote the set of gas samples that are valid at both positions. The gas concentration change is computed as:
\[
\Delta c_t =
\frac{1}{|\mathcal{I}_t|}
\sum_{i \in \mathcal{I}_t}
\left(
c_i(p_t) - c_i(p_{t-1})
\right).
\]
If no gas samples are valid at both positions, the gas gradient bonus is not applied.
The distance reward, including the gas gradient bonus, is defined as:
\[
r_{\mathrm{dist}} =
\begin{cases}
+0.9,
& \text{if } d_{\mathrm{curr}} < d_{\mathrm{prev}}
\text{ and } \Delta c_t > \epsilon_c,\\
+0.5,
& \text{if } d_{\mathrm{curr}} < d_{\mathrm{prev}}
\text{ and } \Delta c_t \leq \epsilon_c,\\
-0.25,
& \text{if } d_{\mathrm{curr}} > d_{\mathrm{prev}},\\
0,
& \text{otherwise},
\end{cases}
\]
where $d_{\mathrm{prev}}$ and $d_{\mathrm{curr}}$ denote the geodesic distances to the source before and after the action, respectively, and $\epsilon_c = 10^{-6}$ is a small numerical tolerance. The $+0.9$ reward consists of the base distance reward of $+0.5$ and an additional gas gradient bonus of $+0.4$. Thus, the bonus is given only when the agent moves closer to the source while reaching a location with a higher mean gas concentration. Turning actions and unsuccessful forward actions leave the agent at the same position and therefore receive zero distance reward.
When the agent selects \texttt{stop}, the distance reward is not applied. Instead, the stop reward is:
\[
r_{\mathrm{stop}} =
\begin{cases}
15 + 5 \cdot
\left(
1 - \dfrac{t_{\mathrm{elapsed}}}{T_{\max}}
\right),
& \text{if the agent stops at the source},\\
-2, & \text{otherwise}.
\end{cases}
\]
Here, $t_{\mathrm{elapsed}}$ is the elapsed simulation time since the episode start, and $T_{\max}$ is the available simulation duration from the episode start to the end of the FDS horizon. The time-dependent bonus gives a larger reward when the source is localized earlier.

\vspace{-6pt}
\subsection{Policy Training}
\vspace{-3pt}
Each training episode contains one odor source sampled from a navigable grid location, and the agent is initialized at another navigable location with a random horizontal orientation. We use a distance-based curriculum during training so that the agent first learns short-range source localization before being exposed to longer-range odor-guided navigation. The initial geodesic distance is linearly increased from 1.5 m to 9.0 m over 30M training steps. The 9 m cap defines a common feasible spawn range across scenes of varying sizes to ensure a consistent training and evaluation setup, rather than reflecting an inherent limitation of the policy. We provide an ablation on this distance cap in the Appendix. An episode terminates when the agent issues the \texttt{stop} action or reaches the 600~s FDS simulation horizon.

Our task follows a detection-triggered navigation setting, motivated by practical deployment scenarios in which source localization is initiated after an odor detection event rather than through continuous uninformed search. Specifically, source localization begins only after the gas concentration at the agent's initial position exceeds a predefined threshold. This trigger is part of the task design, as initiating navigation without evidence of a source would otherwise lead to unnecessary exploration. Accordingly, each episode begins at the first timestep at which the gas density exceeds a scenario- and dataset-specific threshold. For each task and dataset, the threshold is set with reference to the mean value corresponding to 3\% of the scene-wise min-max gas density ranges in the training simulations. The threshold determines only when navigation begins and provides no information about the direction or location of the source. Under this episode setup and curriculum, the policy is optimized using Proximal Policy Optimization (PPO)~\cite{schulman2017proximalpolicyoptimizationalgorithms}.

\vspace{-6pt}
\section{Experiments}
\vspace{-3pt}

\subsection{Evaluation Setup and Metrics}\label{sec:eval_setup_and_metrics}
\vspace{-3pt}

For validation and testing, we randomly sample 10 episodes per scene from the held-out validation and test scenes, respectively. Each episode contains one odor source sampled from a navigable grid location, with the agent initialized at another navigable location within geodesic distance $\min(d_{\max}, 9.0~\mathrm{m})$, where $d_{\max}$ is the maximum reachable geodesic distance from the source in the navigation graph. The initial orientation is sampled uniformly at random. We use one random training seed and three independent test seeds, following prior work~\cite{chen2023soundspaces20simulationplatform}, and report the mean and standard deviation across the three test runs.

As our primary evaluation metric, we use \textit{SPL} (Success weighted by Path Length) \cite{anderson2018evaluationembodiednavigationagents}:
\[
\mathrm{\textit{SPL}}
=
S \cdot \frac{\ell}{\max(p, \ell)},
\]
where $S \in \{0,1\}$ indicates whether the agent selects \texttt{stop} at the source, $\ell$ is the shortest geodesic distance from the start position to the source position, and $p$ is the accumulated geodesic distance along the agent's executed trajectory. We also report the \textit{Final Distance to Target}, defined as the geodesic distance between the agent's final position and the source. 

\vspace{-6pt}
\subsection{Baselines}
\vspace{-6pt}

We compare with three rule-based odor-driven methods and one learning-based method. The rule-based baselines cover three source-seeking strategies: greedily following the strongest local odor cue, using moth-inspired plume-tracking behavior, and estimating a temporal local odor gradient.

\vspace{-9pt}
\paragraph{Chemotactic Search.}
Chemotactic Search greedily follows the strongest local directional odor cue by comparing the center reading with directional readings around the agent. This baseline tests whether local concentration ascent is sufficient for visual-olfactory navigation.

\vspace{-9pt}
\paragraph{Biomimetic Search.}
Biomimetic Search is a moth-inspired heuristic using surge, zigzag, and looping behaviors used in odor plume tracking. Inspired by the state-machine design of the Robust Moth-Inspired (RMI) algorithm~\cite{shigaki2023robust}, it is adapted to our discrete action setting and tests whether classical bio-inspired plume-tracking  transfers to indoor visual-olfactory navigation.

\vspace{-9pt}
\paragraph{Temporal Gradient Search.}
Temporal Gradient Search estimates a local odor gradient from recent gas measurements and agent poses, then moves toward increasing concentration. It is motivated by robotic olfaction methods that infer spatial source direction from temporal chemical signals~\cite{liu2026scensoryrealtimeroboticolfactory}.

\vspace{-9pt}
\paragraph{Dual-Recurrent Threshold RL.}
Dual-Recurrent Threshold Reinforcement Learning (DRT-RL) is a learning-based baseline adapted from prior work~\cite{10473576}. It employs separate recurrent memories for the actor and critic, along with a gas-concentration-based threshold reward. Using the same visual, olfactory, and pose inputs as our policy, it tests whether an alternative recurrent architecture and odor-driven reward design can improve indoor visual-olfactory navigation.

\begin{table*}[t]
\centering
\vspace{0.05em}
\footnotesize
\caption{\textbf{\textcolor{black}{Propane Gas Leak Evaluation Results.}}
SPL in percent; final distance in meters.}

\label{tab:propane_gas_leak_main_results}
\resizebox{0.85\textwidth}{!}{
\begin{tabular}{
l
*{4}{>{\centering\arraybackslash}p{2.1cm}}
}
\toprule
\multirow{2}{*}{Method}
& \multicolumn{2}{c}{Replica}
& \multicolumn{2}{c}{Matterport3D} \\
\cmidrule(lr){2-3}
\cmidrule(lr){4-5}
& \shortstack{SPL \\ $(\%, \uparrow)$}
& \shortstack{Final Dist. \\ $(\mathrm{m}, \downarrow)$}
& \shortstack{SPL \\ $(\%, \uparrow)$}
& \shortstack{Final Dist. \\ $(\mathrm{m}, \downarrow)$} \\
\midrule

Temporal Gradient Search
& $4.59 \pm 0.44$ & $6.28 \pm 0.16$
& $1.83 \pm 0.96$ & $7.79 \pm 0.11$ \\

Biomimetic Search (Moth)
& $36.11 \pm 0.51$ & $4.61 \pm 0.25$
& $33.56 \pm 1.69$ & $4.14 \pm 0.37$ \\

Chemotactic Search
& $17.33 \pm 2.50$ & $5.45 \pm 0.22$
& $27.10 \pm 2.70$ & $4.70 \pm 0.39$ \\

Dual-Recurrent Threshold RL
& $54.50 \pm 5.50$ & $2.03 \pm 1.03$
& ${\mathbf{65.41 \pm 5.15}}$ & $\mathbf{1.54 \pm 0.38}$ \\

\textbf{Ours}
& $\mathbf{65.26 \pm 6.59}$ & $\mathbf{1.48 \pm 0.47}$
& $63.82 \pm 7.85$ & ${1.59 \pm 0.67}$ \\

\bottomrule
\end{tabular}
}
\vspace{-3pt}
\end{table*}

\begin{table*}[t]
\centering
\footnotesize
\caption{\textbf{\textcolor{black}{Methanol Pool Fire Evaluation Results.}}
SPL in percent; final distance in meters.}
\label{tab:methanol_pool_fire_main_results}
\resizebox{0.85\textwidth}{!}{
\begin{tabular}{
l
*{4}{>{\centering\arraybackslash}p{2.1cm}}
}
\toprule
\multirow{2}{*}{Method}
& \multicolumn{2}{c}{Replica}
& \multicolumn{2}{c}{Matterport3D} \\
\cmidrule(lr){2-3}
\cmidrule(lr){4-5}
& \shortstack{SPL \\ $(\%, \uparrow)$}
& \shortstack{Final Dist. \\ $(\mathrm{m}, \downarrow)$}
& \shortstack{SPL \\ $(\%, \uparrow)$}
& \shortstack{Final Dist. \\ $(\mathrm{m}, \downarrow)$} \\
\midrule

Temporal Gradient Search
& $0.00 \pm 0.00$ & $7.27 \pm 0.15$
& $0.00 \pm 0.00$ & $8.73 \pm 0.13$ \\

Biomimetic Search (Moth)
& $0.00 \pm 0.00$ & $7.32 \pm 0.24$
& $0.00 \pm 0.00$ & $8.75 \pm 0.04$ \\

Chemotactic Search
& $0.00 \pm 0.00$ & $7.25 \pm 0.04$
& $0.00 \pm 0.00$ & $8.89 \pm 0.03$ \\

Dual-Recurrent Threshold RL
& $17.58 \pm 3.45$ & $5.07 \pm 0.49$
& $2.74 \pm 1.83$ & $7.60 \pm 0.13$ \\

\textbf{Ours}
& $\mathbf{24.72 \pm 3.69}$ & $\mathbf{4.62 \pm 0.41}$
& $\mathbf{14.16 \pm 2.96}$ & $\mathbf{4.57 \pm 0.65}$ \\

\bottomrule
\end{tabular}
}
\vspace{-6pt}
\footnotesize
\end{table*}

\vspace{-3pt}
\subsection{Experimental Results}
\vspace{-3pt}
Tables~\ref{tab:propane_gas_leak_main_results} and~\ref{tab:methanol_pool_fire_main_results} summarize the results across the two odor-source scenarios and datasets. Both our method and the learning-based baseline consistently outperform the rule-based baselines in both metrics of SPL and final distance to target, demonstrating the benefit of learning from visual geometry, olfactory measurements, and spatio-temporal navigation history. The methanol pool fire scenario is substantially more challenging:
heat-driven buoyancy can cause CO$_2$ cues to rise, spread, or descend depending on scene geometry, producing local concentration peaks that may not reliably indicate the source. As a result, rule-based baselines that rely mainly on local odor cues fail on both datasets. DRT-RL also exhibits limited performance under these conditions. This difference may partly stem from the reward design: DRT-RL's concentration-threshold reward can reinforce misleading odor cues, whereas our distance-based reward shaping encourages the policy to implicitly learn source-seeking behavior. Our policy achieves the highest mean SPL and lowest mean final distance on both datasets. In contrast, propane tends to remain near the floor, producing a more usable concentration structure for reactive search; Consequently, the rule-based baselines perform substantially better than in the fire scenario, while both learning-based methods achieve higher SPL and lower final distances. In the propane gas leak scenario, we achieve higher mean SPL and lower mean final distance than DRT-RL on Replica, whereas DRT-RL shows a slight numerical advantage on Matterport3D. Overall, our policy maintains competitive source-localization performance across both odor-source scenarios with distinct plume characteristics.

\begin{figure*}[!t]
    \centering
    \includegraphics[width=\textwidth]{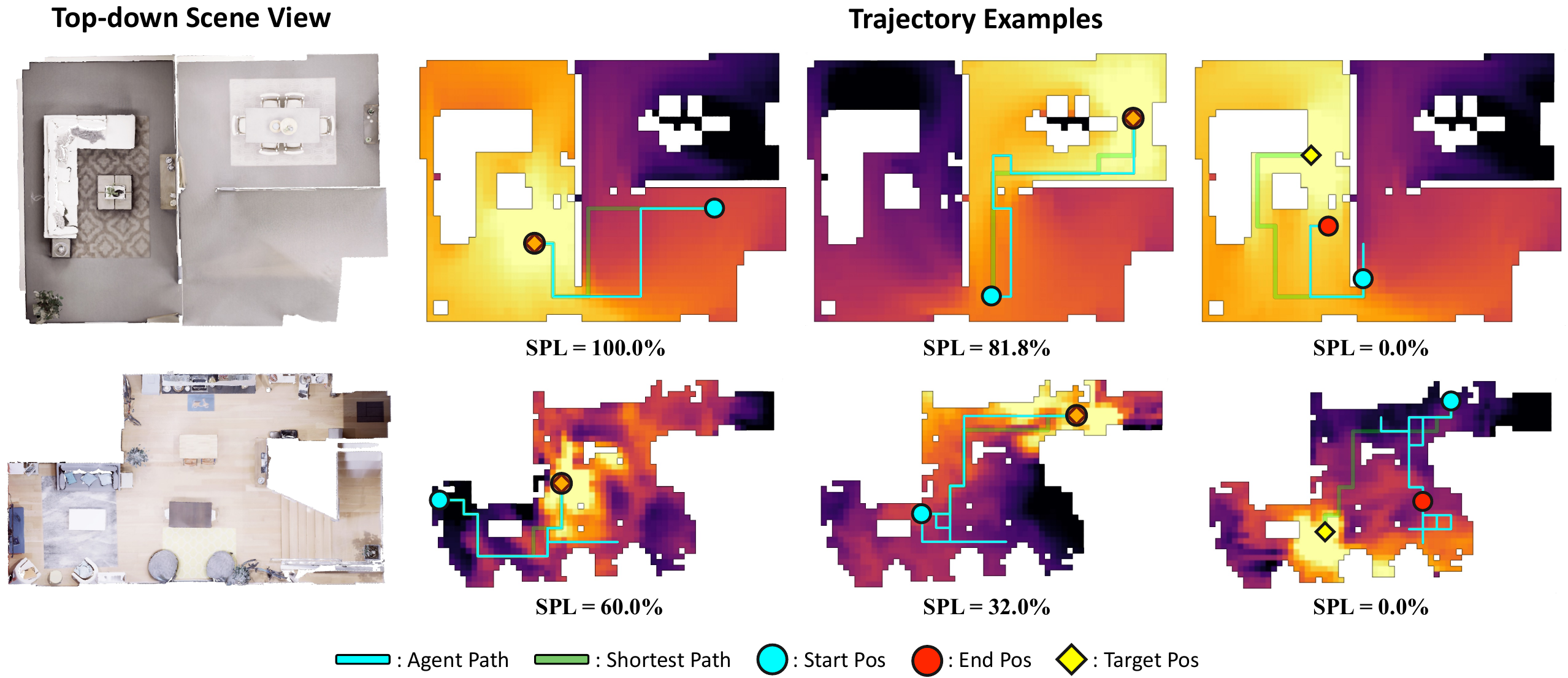}
    \caption{\textbf{Navigation trajectory examples.} The first row shows propane gas leak episodes, and the second row shows methanol pool fire episodes. The first two columns of trajectory examples are success trajectories, where the last column shows representative failure cases.
    }
    \label{fig:trajectory_examples}
    \vspace{-12pt}
\end{figure*}

\textbf{Navigation Trajectory Visualization.} Figure~\ref{fig:trajectory_examples} shows representative successful (first two columns) and failed (last column) trajectories of our policy. 
In the propane gas leak scenario (first row), successful episodes show that the agent often integrates near-floor odor gradients with visual cues to approach the target efficiently. However, failures occur when odor gradients conflict with scene geometry and navigability. In the failure case, the agent follows the odor cue toward a non-navigable obstacle and times out rather than detouring around it. The fire case (second row) reveals a different limitation: heat-driven buoyancy generates noisy plume structures with multiple local concentration peaks, which can attract the agent to a locally strong but non-informative odor region.

\begin{wraptable}{r}{0.48\textwidth}
\vspace{-12pt}
\centering
\small
\caption{\textbf{\textcolor{black}{Input modality ablation.}} We compare ablated variants of our policy using odor (O), vision (V), pose (P), and their combinations as inputs for methanol pool fire in Replica.
}
\label{tab:methanol_input_ablation}
\vspace{0.35em}
\begin{tabular}{lcc}
\toprule
Method & SPL (\%) $\uparrow$ & Final Dist. (m) $\downarrow$ \\
\midrule
O & $14.55 \pm 3.53$ & $5.05 \pm 0.20$ \\
O + V & $19.14 \pm 2.78$ & $4.80 \pm 0.25$ \\
O + P & $16.49 \pm 3.34$ & $5.22 \pm 0.24$ \\
\textbf{O + V + P} & $\mathbf{24.72 \pm 3.69}$ & $\mathbf{4.62 \pm 0.41}$ \\
\bottomrule
\end{tabular}
\vspace{-8pt}
\end{wraptable}

\textbf{Ablation Study.} Table~\ref{tab:methanol_input_ablation} reports an input-modality ablation for the methanol pool fire scenario in Replica. Odor measurements alone yield limited performance, pose history improves SPL, while incorporating vision leads to gains in both SPL and final distance, suggesting that geometric cues help the agent navigate more effectively when odor cues are ambiguous. The full model achieves the best SPL and final distance, showing that all three observation streams contribute to fire-driven odor navigation. See Appendix for more ablation results.

\vspace{-6pt}
\section{Conclusion}
\vspace{-3pt}

We presented \emph{visual-olfactory embodied navigation}, a new task where agents use visual observations and olfactory measurements to navigate toward odor sources in 3D environments, and \emph{OdorSpaces}, a physics-based embodied simulation platform and benchmark for this task. OdorSpaces provides standardized evaluation protocols for two representative scenarios: invisible gas leak localization and early-stage fire localization. We also developed a multimodal reinforcement learning policy that integrates visual and olfactory observations over time and demonstrates competitive source-localization performance against diverse baselines across both scenarios. We hope OdorSpaces, as the first simulation benchmark for visual-olfactory embodied navigation, spurs further research on the role of machine olfaction in embodied navigation.

\vspace{-3pt} \textbf{Limitations and Future Work.} OdorSpaces currently uses precomputed odor fields and idealized olfactory sensing, abstracting away robot-induced airflow, ventilation, and practical sensor constraints such as placement, noise, and response latency~\cite{nose, nosereview, PANDA2024117616, s120709635}. The benchmark also focuses on source localization after odor detection, rather than the full plume-finding problem. Future work can extend OdorSpaces with dynamic airflow modeling that accounts for robot motion, ventilation, and wind, compact and realistic sensor configurations, pre-detection plume finding, and broader odor sources, supporting more robust sim-to-real navigation policies.

\clearpage
\acknowledgments{Research reported in this publication was in part supported by NSF:RITEL 2507056. Any opinions, findings, and conclusions or recommendations expressed in this material are those of the author(s) and do not necessarily reflect the views of the National Science Foundation. We thank Seungjae Lee, Sungbin Oh, members of UMD Multisensory Machine Intelligence Lab, and the anonymous reviewers for their constructive feedback.}

\bibliography{references}

\clearpage
\section{Appendix}

\subsection{FDS Simulation Setup}
We used \texttt{FDS-6.10.1-0-g12efa16-release} for all FDS simulations. Each simulation uses an enclosed computational domain spatially aligned with the corresponding Habitat-Sim environment. All scene boundaries are prescribed as \texttt{ADIABATIC} walls, so that gas transport is isolated to the effects of the source and the scene geometry rather than heat exchange with the environment. We do not impose external wind or mechanical ventilation in the current benchmark; transport is therefore driven entirely by source-induced flow, buoyancy, density differences, molecular and turbulent diffusion, and mixing. The extension to explicit ventilation and wind-driven indoor flows is left to future work and discussed as a limitation.

Each FDS domain is discretized on a uniform Cartesian grid with a cell size of $0.2\,\text{m} \times 0.2\,\text{m} \times 0.2\,\text{m}$, which balances spatial fidelity against the computational cost of generating a large benchmark. The resulting odor field is stored as a four-dimensional tensor of shape $X \times Y \times Z \times T$, where $X$, $Y$, and $Z$ index the spatial grid and $T$ indexes simulation time. Slice outputs are written at 4\,Hz, yielding a temporal resolution of 0.25\,s.

In our FDS simulations, each source location is represented by a $0.2\,\text{m} \times 0.2\,\text{m} \times 0.2\,\text{m}$ cubic odor source object centered at the corresponding navigation grid point. The cube serves both as the navigation target and as the geometric reference for the scenario-specific odor source configuration. For both scenarios, the odor field is simulated for 600~s.

\textbf{Propane Gas Leak.} For each source location, propane is released horizontally from the lateral faces of the cubic source object using \texttt{VOLUME\_FLOW}${}=-4.0\times10^{-3}\,\mathrm{m^{3}/s}$, where the negative sign follows the FDS convention denoting outflow from the source surface. The release is modeled as pure propane (\texttt{MASS\_FRACTION}${}=1.0$) injected at ambient temperature (\texttt{TMP\_FRONT}${}=20\,^{\circ}\mathrm{C}$), with no combustion reaction defined; the simulation captures steady-state dense-gas dispersion only. The numerical setup follows FDS dispersion-modeling conventions from the \texttt{LNG\_Dispersion} validation suite \cite{mcgrattan2013fdsvalidation}, adapted here for a propane release.

\textbf{Methanol Pool Fire.} For each source location, a $0.1\,\text{m}$-deep methanol pool is placed below the upward-facing face of the cubic source object, with ignition provided by four small cylindrical pilots above the pool surface. The methanol pool itself burns via a
pyrolysis model rather than a prescribed heat release rate (HRR): the liquid methanol material is parameterized with heat of vaporization $1{,}099\,\mathrm{kJ/kg}$, boiling temperature
$64.65\,^{\circ}\mathrm{C}$, and density $796\,\mathrm{kg/m^{3}}$, and the evaporated fuel reacts via a single-step methanol combustion reaction (with carbon monoxide (CO) and soot yields each set to $10^{-3}$, reflecting methanol's clean burning profile). The overall setup is adapted from the official FDS validation cases \cite{mcgrattan2013fdsvalidation}.  

\subsection{Scene Processing and Benchmark Generation}
\textbf{Navigation Domain Construction.}
Each Habitat-Sim scene is discretized into a 2D navigation grid with 0.5~m spacing, with valid grid points serving as possible agent positions and candidate odor-source locations. For both Replica and Matterport3D, we use a single floor per scene. For Replica, we estimate the floor height from navmesh-snapped positions, fix all grid points to this height, and retain the largest connected navigable component. We exclude one Replica scene (\texttt{office\_1}) with fewer than 10 valid grid points to avoid trivially short-range episodes. For Matterport3D, which contains larger and often multi-floor environments, we evaluate candidate regions across different floor levels and select a connected subregion based on navigability and geometric structure. This selection is primarily motivated by the computational and storage costs associated with large simulation domains. Using entire Matterport3D scenes would substantially increase both the volume of each computational domain and the number of valid candidate source positions, leading to significantly larger simulation outputs and greater overall computational costs.

\textbf{Scene Geometry Preprocessing.}
Most manual intervention is concentrated in the initial geometry preprocessing stage. To ensure consistency between the visual environment, scene navigability, and simulated gas dynamics, we preprocess scene geometries for both datasets. Specifically, we repair Replica scene meshes to obtain watertight geometry. For Matterport3D, we construct enclosed computational domains by adding ceiling, floor, and wall boundaries, as the selected subregions often lack complete enclosing surfaces after cropping. Without such preprocessing, geometric openings or gaps in the rendered scene may suggest possible gas transport pathways that are not represented in the FDS simulation. For example, a visually open ceiling may lead an agent to expect gas to disperse upward, whereas the corresponding boundary may be treated as a solid surface in FDS, resulting in different gas transport behavior. Such discrepancies could introduce misleading cues when an agent reasons about odor observations in relation to its surrounding geometry.

\textbf{Simulation Generation Pipeline.}
Once the scene geometry is prepared, the processed PLY meshes are converted to GLB format and subsequently transformed into an FDS-compatible block-based representation using \texttt{PyroSim 2024.2.1209}. Based on this representation, FDS template files are created for each scene by incorporating task-specific simulation configurations for propane gas leak and methanol pool fire scenarios. For each candidate source position, a separate FDS input file is then generated from the corresponding scene-specific template. Using these FDS input files, the simulations are executed independently in parallel, and their resulting fields are stored for subsequent use in Habitat-Sim.

\textbf{Dataset Scale and Computational Cost.} For each scenario, we generate 1,415 FDS simulations in Replica and 4,306 in Matterport3D, yielding 5,721 simulations per scenario and 11,442 simulations across both scenarios. For each scene, we allocate approximately 8-16 CPU cores on a compute cluster to run the FDS simulations. No GPUs are used during simulation generation. Depending on the number of candidate source positions and the computational domain size, generating all simulations for a single scene can take up to approximately 2.5 days. All simulations are precomputed offline, and Habitat-Sim queries the stored fields at runtime without executing additional CFD simulations.

\subsection{Propane Density Correction at Ground Level}
In our setup, the gas leak source is modeled as a shapeless and non-visible target for the navigation task, but its FDS implementation uses a finite source box. This creates a sampling artifact: FDS reports zero propane density inside the emitting box because the box is represented as a solid obstruction. Therefore, when the ground-level density lookup at the source position returns zero, we replace it with the maximum propane density among the eight horizontally adjacent cells in the same ground-level layer.

More specifically, the gas leak source is implemented as a single 0.2 m solid obstruction (OBST) sitting on the mesh floor, with four lateral faces emitting propane at a prescribed volumetric flow rate. Because the obstruction is solid, FDS carries no gas-phase data inside it and reports zero propane density there, even though the surrounding fluid cells carry physically meaningful values. The artifact specifically affects propane because propane is denser than air and accumulates near the floor, exactly where ground-level sampling occurs. Fire simulations are unaffected because combustion products are buoyant and rise away from the source, and because fire sensors are placed 0.3 m above the obstruction top and well clear of it. Because the obstruction occupies only one cell and a device's nominal location may snap into an adjacent fluid cell after mesh rounding, the correction is applied only when the sampled ground-level value is exactly zero, taken as evidence that the device landed inside the obstruction rather than next to it.

\subsection{Sensing and Action Setup Details}
\textbf{Vision Sensor.}
Visual observations are provided by an egocentric pinhole depth camera mounted on the agent at a height of 0.5~m. The camera captures depth images at a resolution of $128 \times 128$ pixels with a horizontal field of view of $90^\circ$, providing geometric information about the surrounding environment, including nearby obstacles and navigable space.

\textbf{Olfactory Sensor.} For spatial sensing, gas densities are sampled using a multi-height, five-point layout around the agent. The five horizontal locations are center, forward, backward, left, and right, with directional sensors placed at the end of a 0.5~m sensing arm. If a sensing arm is blocked by an obstacle or exits the environment, the corresponding gas-density reading is set to zero. The vertical sensing heights are scenario-specific. For methanol pool fire task, we sample at 0.5, 1.0, 1.5, and 2.0~m to capture the buoyancy-driven vertical structure of carbon dioxide plumes. For propane gas leak case, we sample at 0, 0.5, 1.0, and 1.5~m to capture near-floor propane density field. These heights keep the sensing layout within a near-ground range compatible with a wheeled ground robot embodiment, with a maximum sensing height of 2.0~m. The readings are ordered consistently to form a fixed-dimensional olfactory observation vector.

For temporal sensing, gas densities are sampled over a scenario-specific interval before each policy decision. For the fire scenario, we use a shorter interval of 4 FDS timesteps (1.0~s), as the heat-driven plume transports the carbon dioxide cue upward after it is generated at the source, forming a concentration field that may be less consistently aligned with the agent's local position. Therefore, we use this shorter interval to base decisions on more recent local carbon dioxide observations. For the gas-leak scenario, we use a longer interval of 10 FDS timesteps (2.5~s) to incorporate propane readings over a wider local temporal window before each decision.

\textbf{Pose Sensor.}
The agent also receives pose-related observations from GPS and compass sensors. GPS provides a two-dimensional position relative to the start location, expressed in grid coordinates on the 0.5 m navigation grid. Since Habitat-Sim uses the $y$-axis as the vertical axis, planar navigation is represented on the $x$-$z$ plane. The compass provides a two-dimensional heading vector for the four grid-aligned orientations, $-z$, $+x$, $+z$, and $-x$. We store these four pose values over the most recent $n=10$ decision steps to provide recent motion context to the policy.

\textbf{Agent Embodiment and Action Space.}
For embodied navigation, the agent is modeled as a cylindrical body with radius $0.2$~m and height $1.5$~m. It moves on a 0.5~m navigation grid, as commonly adopted in prior embodied navigation work~\cite{ chen2020soundspaces, zhu2016targetdrivenvisualnavigationindoor}. The discrete action space is $\mathcal{A}=\{\texttt{move\_forward}, \texttt{turn\_left}, \texttt{turn\_right}, \texttt{stop}\}$. A forward action moves the agent by one grid cell, corresponding to 0.5~m, and turn actions rotate it by $90^\circ$. We set the forward speed to 0.2~m/s and the angular speed to $\pi/5~\mathrm{rad/s}$, making all non-stop actions take the same duration of 2.5~s. Since FDS outputs are stored at 4~Hz, each non-stop action advances the odor field by 10 FDS timesteps.

\subsection{Evaluation Setup and Baselines Details}
For all rule-based baselines, multi-height, multi-positions and temporal gas readings are summarized using either mean- or max-based aggregation, and we report the better result between the two variants. 

\textbf{Chemotactic Search.}
The chemotactic baseline follows the strongest local directional odor cue using a five-point olfactory sensor layout consisting of center, front, left, right, and back measurements. At each step, the agent compares the four directional gas readings and selects the direction with the highest concentration. If the highest reading is in front, the agent moves forward; if it is on the left or right, the agent turns in that direction. If the back reading is highest, the agent turns toward the side with the higher lateral concentration. The agent stops when the center reading is greater than all four directional readings. Following a collision, it instead selects a random left or right turn.

\textbf{Biomimetic Search.}
At each step, the agent compares the front and center gas readings. When the front reading exceeds the center reading, the agent executes a forward action (surge) and resets its search state. Consecutive forward actions are performed as long as this condition remains satisfied. Otherwise, the agent enters a zigzag state, alternating left and right turns from a randomly selected initial direction. After three zigzag turns without a higher front reading, the agent transitions to a looping state, repeatedly turning in a fixed direction. From either the zigzag or looping state, a higher front reading triggers a return to surge behavior. If a collision occurs during forward movement, the agent resets to the zigzag state and turns in its stored turning direction. The agent issues a stop action when the center gas reading is greater than all four surrounding readings.

\textbf{Temporal Gradient Search.}
To match the pose-history length used by our policy, the temporal gradient baseline maintains up to $n=10$ most recently updated samples from distinct grid locations, where each sample contains the position relative to the starting location $(x_i,z_i)$, elapsed time $t_i$, and center gas reading $c_i$. When the agent revisits a grid location, the corresponding sample is updated with the latest time and gas measurement. Given these samples, the baseline fits a spatiotemporal linear model
\[
c_i \approx \alpha x_i + \beta z_i + \gamma t_i + b.
\]
Before fitting, the position and time variables are standardized using the statistics of the stored samples. The regression is performed only when at least four samples are available and the full design matrix has rank four; otherwise, the agent follows a simple exploration fallback. The estimated spatial coefficients are transformed back to the original coordinate units, and the resulting gradient $(\alpha,\beta)$ is normalized to determine the navigation direction. The agent selects the forward action when its heading has the highest alignment with the estimated gradient; otherwise, it turns toward the side with greater alignment. The agent issues a stop action when the center gas reading is greater than all four surrounding readings, taking precedence over gradient-based navigation. This baseline evaluates whether recent position, time, and gas observations can support spatial odor-gradient ascent without learning a navigation policy.

\textbf{Dual-Recurrent Threshold RL.}
The DRT-RL baseline uses the same visual, olfactory, and pose-history observations as our policy. Its actor and critic each employ a two-branch architecture consisting of a feedforward branch for the current observation and a recurrent branch for historical information. The recurrent branches maintain independent LSTM \cite{10.1162/neco.1997.9.8.1735} memories for the actor and critic, each processing the previous encoded observation and previous action through a separate history encoder. The current-observation and recurrent features are then fused to produce the actor and critic representations. The actor predicts a categorical distribution over four discrete navigation actions, while the critic estimates the state value. We adopt the dual-recurrent architecture from the original work but train with PPO, as we do for our policy. The reward function also builds on the original formulation, with modifications for our setting. Specifically, the agent receives a reward of $+1$ only when visiting a grid location for the first time and at least one valid gas sensor reading meets or exceeds a predefined concentration threshold. We set this threshold to the same value as the episode-start threshold used in our policy. All other non-terminal steps receive zero reward. A successful stop at the odor source yields $+100$, whereas an unsuccessful stop receives zero reward and a timeout incurs a penalty of $-100$.

\subsection{Ablation and Analysis}
We further analyze the effects of input modalities, curriculum learning, reward design, spatiotemporal olfactory sensing, and recurrent memory in Replica.
Across both odor-source scenarios, the distance-based curriculum and reward shaping improve SPL, with the distance reward complementing the terminal stop reward by providing intermediate progress feedback. Since individual gas sensors provide only local concentration measurements without inherent directional information, sensing across multiple locations, heights, and timesteps is important for capturing the spatiotemporal structure of odor plumes. The gas-gradient bonus further encourages the policy to identify and follow plume structures, making richer temporal, horizontal, and vertical sensing particularly valuable. Consistent with this, multiple timesteps, horizontally distributed sensing locations, and multiple sensing heights yield higher SPL than their corresponding reduced configurations. Recurrent memory, however, exhibits task-dependent effects, improving performance in the propane gas leak scenario but yielding lower performance in the methanol pool fire scenario.

\textbf{Propane Gas Leak.} Propane is denser than air and tends to accumulate and spread along the floor, producing an informative concentration structure for source seeking. The input-modality ablation shows that adding either vision or pose history to olfactory observations improves SPL, with their combination achieving the highest performance among the evaluated input configurations. Multiple sensing heights also improve performance, suggesting that observations at different elevations help the policy capture the vertical concentration structure of the plume. For the single-height ablation, olfactory sensing is fixed at ground level (0 m) to retain the near-floor concentration cues.

\begin{table}[!htb]
\vspace{-10pt}
\centering
\small
\caption{\textbf{Propane gas leak ablations.} We evaluate the effects of input modalities (odor (O), vision (V), and pose history (P)), distance-based curriculum, reward components, temporal and spatial olfactory sensing, and recurrent memory in the Replica propane gas leak scenario.}
\vspace{0.35em}
\begin{tabular}{llcc}
\toprule
Ablation & Method & SPL (\%) $\uparrow$ & Final Dist. (m) $\downarrow$ \\
\midrule
\multirow{4}{*}{Input}
& O & $49.47 \pm 2.99$ & $2.88 \pm 0.29$ \\
& O + V & $58.31 \pm 3.18$ & $2.52 \pm 0.53$ \\
& O + P & ${60.64 \pm 5.84}$ & ${1.98 \pm 0.24}$ \\
& \textbf{O + V + P} & $\mathbf{65.26 \pm 6.59}$ & $\mathbf{1.48 \pm 0.47}$ \\
\midrule
\multirow{2}{*}{Curriculum}
& Without Curriculum & ${53.57 \pm 5.49}$ & ${2.54 \pm 0.31}$ \\
& \textbf{With Curriculum} & $\mathbf{65.26 \pm 6.59}$ & $\mathbf{1.48 \pm 0.47}$ \\
\midrule
\multirow{2}{*}{Reward}
& Stop Reward Only & $51.56 \pm 6.10$ & $3.06 \pm 0.56$ \\
& \textbf{Stop Reward + Distance Reward} & $\mathbf{65.26 \pm 6.59}$ & $\mathbf{1.48 \pm 0.47}$ \\

\midrule
\multirow{2}{*}{Temporal}
& Single Timestep
& $61.48 \pm 6.67$ & $1.78 \pm 0.74$ \\
& \textbf{Multiple Timesteps}
& $\mathbf{65.26 \pm 6.59}$ & $\mathbf{1.48 \pm 0.47}$ \\

\midrule
\multirow{2}{*}{Spatial (Horizontal)}
& Center Only
& $12.88 \pm 1.72$ & $3.79 \pm 0.30$ \\
& \textbf{Gas Grid (Center + 4 directions)}
& $\mathbf{65.26 \pm 6.59}$ & $\mathbf{1.48 \pm 0.47}$ \\

\midrule
\multirow{2}{*}{Spatial (Vertical)}
& Single Height
& $52.62 \pm 10.25$ & $2.64 \pm 0.53$ \\
& \textbf{Multiple Heights}
& $\mathbf{65.26 \pm 6.59}$ & $\mathbf{1.48 \pm 0.47}$ \\

\midrule
\multirow{2}{*}{Recurrent Memory}
& Without Recurrent Memory
& $54.00 \pm 4.94$ & $2.58 \pm 0.24$ \\
& \textbf{With Recurrent Memory}
& $\mathbf{65.26 \pm 6.59}$ & $\mathbf{1.48 \pm 0.47}$ \\
\bottomrule
\end{tabular}
\end{table}

\textbf{Methanol Pool Fire.} In contrast, the methanol pool fire scenario is particularly challenging because buoyant transport can produce odor distributions that are weakly aligned with the agent's ground-level motion. The distance-based curriculum helps the agent gradually learn source-seeking behavior before encountering longer-range episodes. For the single-height ablation, olfactory sensing is fixed at the agent's height of 1.5 m above the ground, considering that buoyancy-driven CO$_2$ initially rises toward the ceiling and gradually descends as it disperses throughout the room. The substantial improvement from multiple sensing heights highlights the importance of capturing vertical concentration variations in this scenario.

\begin{table}[!htb]
\vspace{-10pt}
\centering
\small
\caption{\textbf{Methanol pool fire ablations.} We evaluate the effects of distance-based curriculum, reward components, temporal and spatial olfactory sensing, and recurrent memory in the Replica methanol pool fire scenario.}
\label{tab:ablation_methanol_replica_curriculum_reward_combined}
\vspace{0.35em}
\begin{tabular}{llcc}
\toprule
Ablation & Method & SPL (\%) $\uparrow$ & Final Dist. (m) $\downarrow$ \\
\midrule
\multirow{2}{*}{Curriculum}
& Without Curriculum & $20.43 \pm 6.73$ & $4.85 \pm 0.50$ \\
& \textbf{With Curriculum} & $\mathbf{24.72 \pm 3.69}$ & $\mathbf{4.62 \pm 0.41}$ \\
\midrule
\multirow{2}{*}{Reward}
& Stop Reward Only & $0.00 \pm 0.00$ & $6.17 \pm 0.49$ \\
& \textbf{Stop Reward + Distance Reward} & $\mathbf{24.72 \pm 3.69}$ & $\mathbf{4.62 \pm 0.41}$ \\
\midrule
\multirow{2}{*}{Temporal}
& Single Timestep
& $21.95 \pm 0.49$ & $\mathbf{4.33 \pm 0.29}$ \\
& \textbf{Multiple Timesteps}
& $\mathbf{24.72 \pm 3.69}$ & ${4.62 \pm 0.41}$ \\

\midrule
\multirow{2}{*}{Spatial (Horizontal)}
& Center Only
& $16.98 \pm 1.26$ & $\mathbf{4.04 \pm 0.17}$ \\
& \textbf{Gas Grid (Center + 4 directions)}
& $\mathbf{24.72 \pm 3.69}$ & ${4.62 \pm 0.41}$ \\

\midrule
\multirow{2}{*}{Spatial (Vertical)}
& Single Height
& $0.00 \pm 0.00$ & $5.06 \pm 0.36$ \\
& \textbf{Multiple Heights}
& $\mathbf{24.72 \pm 3.69}$ & $\mathbf{4.62 \pm 0.41}$ \\

\midrule
\multirow{2}{*}{Recurrent Memory}
& \textbf{Without Recurrent Memory}
& $\mathbf{28.69 \pm 4.23}$ & $\mathbf{3.73 \pm 0.08}$ \\
& With Recurrent Memory
& $24.72 \pm 3.69$ & ${4.62 \pm 0.41}$ \\
\bottomrule
\end{tabular}
\end{table}

\FloatBarrier

\subsection{Generalization to Longer Spawn Distances} The 9 m spawn-distance cap defines a common feasible spawn range across scenes of varying sizes for consistent benchmark evaluation, rather than an inherent limitation of the policy. To examine generalization beyond this range, we conduct a controlled ablation using the methanol pool fire scenario in the Replica \texttt{apartment\_0} scene. We train policies with maximum source distances of 7, 9, and 11 m and evaluate them at distances ranging from 5 to 20 m.

The results, measured in SPL (\%), show that extending the training distance range generally improves performance, particularly at longer test distances. Policies trained with broader distance ranges achieve better performance on distant targets, including those beyond the maximum training distance. Although performance declines at the longest test distances, the improvements from longer-range training suggest that the policy can learn source-seeking behavior that generalizes beyond the distances encountered during training.

\begin{table}[!htb]
\vspace{-10pt}
\centering
\small
\caption{\textbf{Generalization to longer spawn distances.} SPL (\%) for different training and test spawn distances in the Replica \texttt{apartment\_0} methanol pool fire scenario.}
\vspace{0.35em}
\resizebox{\columnwidth}{!}{%
\begin{tabular}{lccccccc}
\toprule
& \multicolumn{7}{c}{Test Distance (m)} \\
\cmidrule(lr){2-8}
Train Dist. & 5 m & 7 m & 9 m & 11 m & 13 m & 15 m & 20 m \\
\midrule
7 m & ${57.26 \pm 5.86}$ & $63.00 \pm 0.34$ & $58.77 \pm 8.80$ & $45.13 \pm 5.23$ & $36.52 \pm 5.77$ & $29.40 \pm 8.02$ & $23.61 \pm 4.12$ \\
9 m & $60.86 \pm 1.55$ & $\mathbf{72.80 \pm 4.18}$ & ${74.37 \pm 2.05}$ & $76.79 \pm 2.66$ & $57.66 \pm 2.85$ & $49.37 \pm 3.17$ & $44.90 \pm 3.22$ \\
11 m & $\mathbf{63.65 \pm 4.24}$ & $71.49 \pm 2.76$ & $\mathbf{80.77 \pm 1.94}$\ & $\mathbf{79.60 \pm 7.11}$ & $\mathbf{79.76 \pm 6.81}$ & $\mathbf{66.77 \pm 5.72}$ & $\mathbf{49.17 \pm 4.42}$\\

\bottomrule
\end{tabular}%
}
\end{table}

\end{document}